\documentclass{article}

\usepackage{spconf,amsmath,amssymb,graphicx}
\usepackage{booktabs} 
\usepackage{multirow}
\usepackage{enumitem}
\usepackage{caption}
\usepackage{subcaption}
\usepackage{tikz}
\usetikzlibrary{shapes.geometric, arrows.meta, positioning, fit, calc, backgrounds}
\usepackage{url}

\title{CSC-MAMBA: Bridging Multimodal Convolutional Sparse Coding and State Space Models for Multi-Contrast MRI Super-Resolution}
\name{Iman Marivani, David Van Hamme, Jan Aelterman,  Wilfried Philips}
\address{TELIN-IPI, Ghent University–Imec, Ghent, Belgium.}
\begin{document}
%
\maketitle

\begin{abstract}
Multi-contrast MRI super-resolution accelerates clinical protocols by reconstructing high-resolution (HR) target modalities from low-resolution (LR) inputs, guided by easily acquired reference contrasts. While purely data-driven deep neural networks achieve strong performance via rich representation capacity, they rely on heuristic fusion mechanisms (e.g., naive concatenation or black-box cross-attention) that lack theoretical grounding. Alternatively, model-based deep unfolding networks enforce principled cross-contrast fusion through explicit priors, yet their unrolled steps remain constrained to shallow convolutions, lacking modern deep blocks to capture global context. To bridge this gap, we propose CSC-MAMBA, unifying deep-unfolded multimodal convolutional sparse coding with the Vision Mamba state-space backbone. By fusing common structural priors from the guidance modality via unfolded sparse coding and globally refining contrast-specific representations with a Mamba backbone, our approach benefits from interpretable, model-guided fusion and long-range spatial context. Experiments on the IXI dataset demonstrate that CSC-MAMBA outperforms state-of-the-art methods in reconstruction quality, achieving up to a 0.62 dB gain in PSNR while remaining computationally efficient.
\end{abstract}




\section{Introduction}
Multi-contrast magnetic resonance imaging (MRI) captures complementary tissue characteristics across sequences: T1-weighted images (T1WI) delineate anatomy, T2-weighted (T2WI) and fluid-attenuated inversion recovery (FLAIR) highlight lesions and edema, and proton-density-weighted images (PDWI) aid cartilage assessment~\cite{lyu2020multi, feng2021multi, feng2025bridging, feng2023multimodal}. However, differing acquisition parameters create substantial disparities in scan times. While T1WI and PDWI can be acquired rapidly at high resolution, slower sequences such as T2WI, FLAIR, and FS-PDWI require prolonged scans that reduce patient comfort and increase susceptibility to motion artifacts~\cite{feng2023multimodal, feng2025bridging}. Consequently, clinical protocols often pair fast HR sequences with accelerated, low-resolution acquisitions of slower contrasts (e.g., via $k$-space truncation)~\cite{feng2025bridging, mao2023disc}. Multi-contrast MRI super-resolution (MCSR) accelerates these clinical protocols by leveraging an easily acquired HR reference (e.g., T1WI) as structural guidance to reconstruct a target HR image from its LR counterpart. 

Existing MCSR methods follow four main categories, yet each faces distinct architectural limitations. (1) \textit{Classical optimization and sparse coding} exploit non-local patch similarity~\cite{qu2014magnetic} and coupled dictionary learning~\cite{weizman2016reference, song2020coupled}. Despite their interpretability, handcrafted priors and iterative solvers limit their flexibility and inference speed. (2) \textit{End-to-end deep architectures}~\cite{lyu2020multi, feng2021multi, feng2023multimodal, zeng2018simultaneous, li2022transformer,  ddh2026mamba} capture rich representations, yet they merge modalities through uninterpretable mechanisms---like feature concatenation or cross-modal attention---lacking explicit structural priors. This unconstrained fusion often causes texture-copying artifacts, where reference features erroneously overwrite target-specific structures~\cite{feng2025bridging, song2020coupled}. (3) \textit{Generative diffusion models}~\cite{mao2023disc, li2024rethinking} recover high-frequency details, yet require costly iterative sampling and risk hallucinating anatomically inconsistent structures. (4) \textit{Deep unfolding networks} bridge model-based optimization and deep learning by unrolling iterative solvers into learnable stages, enforcing principled, interpretable cross-contrast fusion~\cite{yang2022model, lei2023decomposition, shen2026unfolding}. However, because these unrolled steps are mathematically constrained to shallow convolutions with limited receptive fields, they fail to natively inherit modern deep blocks capable of modeling the global context and representations necessary for complex tissue structures.

To address these challenges, we introduce \textbf{CSC-MAMBA}, a novel architecture that integrates the Vision Mamba backbone with a deep-unfolded multimodal convolutional sparse coding (CSC) pipeline. Rather than treating cross-contrast fusion as an unconstrained black box, our method exploits joint sparse modeling to fuse shared anatomical structures from the guidance modality while preserving target-specific details, providing an effective framework for cross-modal guidance. The resulting sparse representations are subsequently refined by Vision Mamba modules, injecting long-range spatial context across the image features. By bridging the interpretability of model-based optimization with the expressive power of global state-space models, CSC-MAMBA achieves superior reconstruction performance on the IXI benchmark without compromising the inference speed.

\section{Related Work}
\label{sec:related}

\noindent\textbf{Data-Driven Multi-Contrast SR:} End-to-end deep learning models directly map LR targets to HR outputs by leveraging cross-contrast fusion, transferring sharp high-frequency anatomical features from an auxiliary HR reference~\cite{lyu2020multi, feng2021multi, feng2025bridging, zeng2018simultaneous}. Despite their empirical success, standard CNNs are constrained by localized receptive fields. While Vision Transformers (McMRSR~\cite{li2022transformer}, MTrans~\cite{feng2023multimodal}) address this, they scale with quadratic computational complexity $\mathcal{O}(N^2)$ relative to sequence length. Generative diffusion models (DisC-Diff~\cite{mao2023disc}) synthesize rich textures but suffer from prolonged, multi-step iterative sampling and risk hallucinating anatomically inaccurate features. Crucially, unconstrained feature fusion in these ``black-box'' models frequently induces texture-copying artifacts, where contrast-specific reference details are erroneously hallucinated into target structures.

\noindent\textbf{Model-Based Priors and Deep Unfolding:} Classical methods formulate reconstruction as an inverse problem, optimizing explicit mathematical objectives subject to structural priors~\cite{qu2014magnetic, weizman2016reference, ehrhardt2016multicontrast}. While mathematically rigorous and interpretable, these iterative optimization algorithms suffer from high computational overhead. Coupled Dictionary Learning~\cite{song2020coupled} mitigates texture copying by explicitly disentangling modality-shared structural priors from contrast-specific details. Deep Unfolding Networks bridge model-based optimization and deep learning by unrolling iterative algorithms---such as multimodal CSC---into learnable neural network layers with principled, interpretable fusion rules (e.g.,~\cite{yang2022model},~\cite{lei2023decomposition},~\cite{shen2026unfolding}). However, conventional deep unfolded networks typically map proximal updates directly to shallow, localized convolutional blocks. Consequently, they fail to capture long-range spatial context or maximize the expressiveness required for complex cross-contrast dependencies.

\noindent\textbf{State-Space Models in MRI Reconstruction:} Structured State-Space Models (Mamba)~\cite{gu2023mamba} capture long-range dependencies with linear $\mathcal{O}(N)$ complexity via selective state scanning. Recent studies have adapted Mamba for MRI reconstruction (MambaMIR~\cite{huang2024mambamir}), deformable single-contrast SR (Deform-Mamba~\cite{ji2024deform}), and multi-contrast SR~\cite{ddh2026mamba}. However, existing Mamba-based MCSR methods still rely on heuristic feature concatenation or black-box attention for cross-contrast fusion. In contrast, our proposed \textbf{CSC-MAMBA} bridges an unfolded multimodal CSC pipeline with the Vision Mamba backbone---combining theoretically explainable cross-contrast fusion with efficient, global spatial modeling.

\begin{figure*}[t!]
  \centering
  \includegraphics[width=1.0\textwidth]{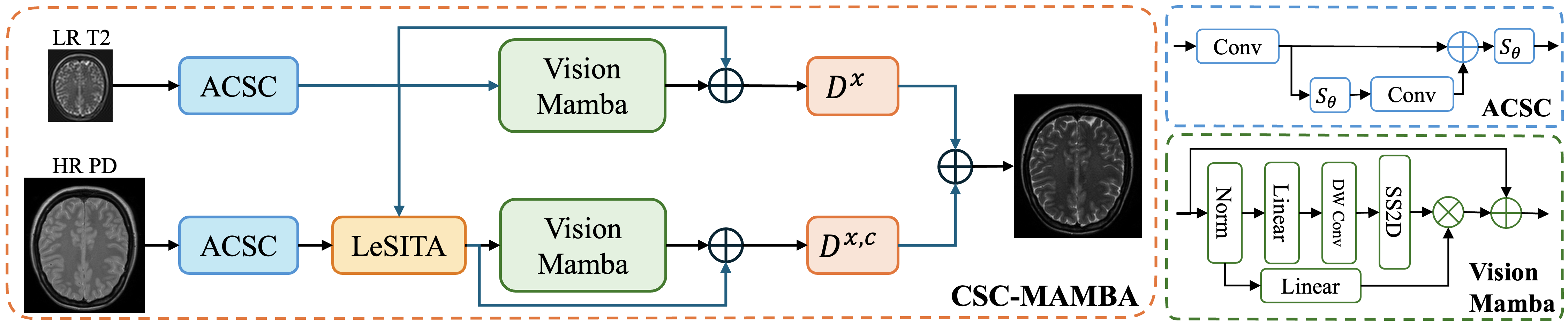}
  
  \caption{Proposed model architecture.}
  \label{fig:model_architecture}
\end{figure*}

\section{The Proposed Model: CSC-MAMBA}
\label{sec:proposed}

In this section, we present our hybrid multi-contrast super-resolution model, \textbf{CSC-MAMBA}, which integrates coupled convolutional sparse coding priors with Vision Mamba state-space backbones. The overall architecture of our model is depicted in Fig.~\ref{fig:model_architecture}.

\subsection{Multimodal Convolutional Sparse Coding with Side Information}
We represent the LR target image~$\boldsymbol{y}$, HR reference guidance~$\boldsymbol{z}$, and target HR image~$\boldsymbol{x}$ using convolutional sparse coding as:
\begin{equation}
\boldsymbol{v} = \sum\limits_{i=1}^{k}\boldsymbol{d}^{v}_i*\boldsymbol{\alpha}^{v}_i, \quad \text{for } v \in \{y, z, x\},
\label{eq:mcsc_decomp}
\end{equation}
where $*$ denotes 2D convolution between dictionary atoms $\boldsymbol{d}^{v}_i$ and convolutional feature maps $\boldsymbol{\alpha}^{v}_i$. Guided by the principle that coupled sparse coding of correlated modalities can be formulated via $\ell_1$-$\ell_1$ minimization~\cite{song2020coupled}, we model cross-contrast fusion as a multimodal convolutional sparse coding (MCSC) problem with side information~\cite{marivani2020multimodal}:
\begin{equation}
\label{eq:MCSC}
\begin{aligned}
\min_{\boldsymbol{\alpha}} \; \frac{1}{2}\| \boldsymbol{y} - \sum\limits_{i=1}^{k}\boldsymbol{d}_i*\boldsymbol{\alpha}_i \|^2_2
 +\lambda\sum\limits_{i=1}^{k}(\|\boldsymbol{\alpha}_i\|_1 +\|\boldsymbol{\alpha}_i - \boldsymbol{\alpha}^{z}_i\|_1).
\end{aligned}
\end{equation} 

This optimization is solved via an unfolded learned multimodal convolutional sparse coding (LMCSC) network~\cite{marivani2020multimodal}. Each recurrent step updates the coupled representation using the Learned Soft-Iterative Thresholding Algorithm (LeSITA) proximal operator~$\xi_{\mu}$:
\begin{equation}
\label{eq:lesita}
\boldsymbol{\alpha}^{\tau+1} = \xi_{\mu}\big(\boldsymbol{\alpha}^\tau - \boldsymbol{Q}*\boldsymbol{R}*\boldsymbol{\alpha}^\tau + \boldsymbol{P}*\boldsymbol{y};\; \boldsymbol{\alpha}^z\big),
\end{equation}
where $\boldsymbol{Q}$, $\boldsymbol{R}$, $\boldsymbol{P}$ are learnable convolutional layers, and threshold~$\mu$ is a learnable parameter. This fused sparse representation encodes the shared structural information between both modalities.

\subsection{Dual-Branch CSC-MAMBA Architecture}
CSC-MAMBA employs a two-branch architecture comprising LR target features and an HR fused feature space derived from multimodal convolutional sparse coding. The resulting convolutional sparse feature maps act as dynamic states, which are updated via resolution-specific Vision Mamba networks to model rich, global intra-modal dependencies.

\noindent
\textbf{Initial Sparse Extraction and LeSITA Fusion:} 
The proposed pipeline begins by processing both the LR target image $\boldsymbol{y}$ and the HR guidance image $\boldsymbol{z}$ through an approximate convolutional sparse coding (ACSC) network~\cite{sreter2018learned} to extract their respective convolutional sparse feature maps (see Fig.~\ref{fig:model_architecture}). Here, $\mathcal{S}_{\boldsymbol{\theta}}(u) = \operatorname{sign}(u)\max(\vert{}u\vert{}-\boldsymbol{\theta},0)$ denotes the soft thresholding operator utilized during extraction. Accordingly, the initial LR state $(t=0)$ is defined as $S^0_l = \mathrm{ACSC}_y(\boldsymbol{y})$.
Moreover, the initial HR state is generated using the LeSITA network. This module fuses the extracted convolutional sparse representation of the guidance image, denoted as $\boldsymbol{\alpha}^z = \mathrm{ACSC}_z(\boldsymbol{z})$, with the sparse representation of the LR state. Formally, this initial HR state is given by:
\begin{equation}
S^0_h = \mathrm{LeSITA}\big(S^0_l;\; \boldsymbol{\alpha}^z\big),
\label{eq:init_hr}
\end{equation}
where $\mathrm{LeSITA}(\cdot)$ is the unfolded LeSITA update given in~\eqref{eq:lesita}. These two sparse feature maps serve as the initial representations for the two branches of the network. Specifically, $S^0_l$ encapsulates the features specific to the target modality, while $S^0_h$ represents the initialized fused state that captures the correlated structural information between the modalities.

\noindent
\textbf{Mamba-Driven Representation Refinement:} 
In the subsequent step, the LR and HR representations are refined by modality-specific Vision Mamba networks, $\Phi_{\mathrm{Mamba}}^{\mathrm{LR}}(\cdot)$ and $\Phi_{\mathrm{Mamba}}^{\mathrm{HR}}(\cdot)$, which capture global long-range spatial dependencies~\cite{zhu2024vision}. The relationships between the states are expressed as: 
\begin{equation}
S^t_l = \gamma\,\mathrm{ACSC}_y(\boldsymbol{y}) + \Phi_{\mathrm{Mamba}}^{\mathrm{LR}}(S^{t-1}_l),
\label{eq:lr_update}
\end{equation} 
\begin{equation}
S^t_h = \Phi_{\mathrm{Mamba}}^{\mathrm{HR}}(S^{t-1}_h) + \mathrm{LeSITA}\big(S^t_l;\; \boldsymbol{\alpha}^z\big),
\label{eq:hr_update}
\end{equation} 
where $\gamma$ controls the influence of the initial LR sparse representation on the current LR state. Specifically, as shown in Fig.~\ref{fig:model_architecture}, for an input state $S \in \mathbb{R}^{H\times W\times C}$, the Vision Mamba block first applies layer normalization and a linear projection before splitting the features into two parallel branches. One branch undergoes a simple linear transformation, while the other is sequentially processed by a linear layer, a depthwise convolution ($\mathrm{DW Conv}$), and a 4-way 2D Selective Scan ($\mathrm{SS2D}$) to capture global spatial dependencies. The outputs of both branches are then aggregated to produce the final refined state~\cite{zhu2024vision}. While this process can be iterated $T$ times for progressive refinement, in this paper, we employ only a single iteration of Vision Mamba blocks to maintain high computational efficiency.

\noindent
\textbf{Reconstruction and Training Objective:} 
Finally, we reconstruct the super-resolved target image $\hat{\boldsymbol{x}}$ by passing the refined sparse feature maps, $S^T_h$ and $S^T_l$ (with $T=1$), through separate convolutional dictionaries~$\boldsymbol{D}^{x,c}$ and~$\boldsymbol{D}^x$, respectively, and summing their outputs: $\hat{\boldsymbol{x}} = \boldsymbol{D}^{x,c} * S^T_h + \boldsymbol{D}^x * S^T_l$. This step explicitly recovers and merges both the correlated structural information and the modality-specific details. We train the network end-to-end by minimizing the Frobenius norm between the ground-truth (GT) image~$\boldsymbol{x}$ and the network output~$\hat{\boldsymbol{x}}$ across training pairs:
\begin{equation}
\min_{\Theta} \sum\limits_{n} \| \boldsymbol{x}_n - \hat{\boldsymbol{x}}_n \|_F^2,
\label{eq:training_loss}
\end{equation}
where~$\Theta$ denotes the set of all learnable network parameters.

\begin{table}[t]
\centering
\caption{Quantitative results and inference times for PD-guided T2 SR ($\times2$ and $\times4$) on the IXI dataset.}
\label{tab:quantitative_comparison}
\small
\setlength{\tabcolsep}{2.5pt} 
\renewcommand{\arraystretch}{1.1}
\resizebox{\linewidth}{!}{
\begin{tabular}{l c c c c c}
\toprule
\multirow{2}{*}{Method} & \multirow{2}{*}{\begin{tabular}{@{}c@{}}Inference \\ Time (ms)\end{tabular}} &
\multicolumn{2}{c}{$\times2$ SR} &
\multicolumn{2}{c}{$\times4$ SR} \\
\cmidrule(lr){3-4} \cmidrule(lr){5-6}
& & PSNR $\uparrow$ & SSIM $\uparrow$
& PSNR $\uparrow$ & SSIM $\uparrow$ \\
\midrule
DDHMamba~\cite{ddh2026mamba}
    & 283 & 40.89 & 0.9776 & 37.29 & 0.9597 \\
McVarNet~\cite{lei2023decomposition}
    & 62 & 41.87 & 0.9798 & 37.89 & 0.9648 \\
McMRSR~\cite{li2022transformer}
    & 970 & 40.37 & 0.9754 & 37.12 & 0.9589\\
DiffMSR~\cite{li2024rethinking}
    & 39 & 41.62 & 0.9812 & 37.78 & 0.9644 \\
CD-DPE~\cite{gu2026cddpe}
    & 98 & 42.08 & 0.9821 & 38.14 & 0.9668 \\
HocMRI~\cite{shen2026unfolding}
    & 112 & 41.97 & 0.9815 & 38.07 & 0.9651 \\
\midrule
\textbf{CSC-MAMBA}
    & 52 & \textbf{42.57} & \textbf{0.9843}
    & \textbf{38.76} & \textbf{0.9681} \\
\bottomrule
\end{tabular}%
}
\end{table}

\begin{figure}[tbp]
    \centering
    \captionsetup[subfigure]{labelformat=empty} 
    \setlength{\tabcolsep}{0pt}
    \setlength{\fboxrule}{0.5pt} 
    \setlength{\fboxsep}{0pt}   
    
    \begin{subfigure}[b]{0.132\linewidth}
        \centering
        \fbox{\includegraphics[width=\linewidth]{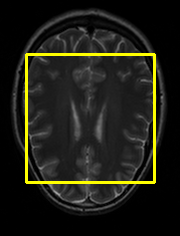}}
        \caption{\small GT}
        \label{fig:hr_box}
    \end{subfigure}%
    \begin{subfigure}[b]{0.173\linewidth}
        \centering
        \fbox{\includegraphics[width=\linewidth]{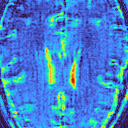}}
        \caption{\small McMRSR}
        \label{fig:diff_swinir}
    \end{subfigure}%
    \begin{subfigure}[b]{0.173\linewidth}
        \centering
        \fbox{\includegraphics[width=\linewidth]{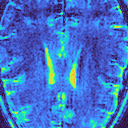}}
        \caption{\small McVarNet}
        \label{fig:diff_varnet}
    \end{subfigure}%
    \begin{subfigure}[b]{0.173\linewidth}
        \centering
        \fbox{\includegraphics[width=\linewidth]{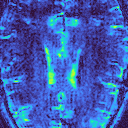}}
        \caption{\small DiffMSR}
        \label{fig:diff_diffmsr}
    \end{subfigure}%
    \begin{subfigure}[b]{0.173\linewidth}
        \centering
        \fbox{\includegraphics[width=\linewidth]{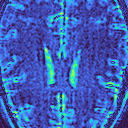}}
        \caption{\small CD-DPE}
        \label{fig:diff_csc}
    \end{subfigure}%
    \begin{subfigure}[b]{0.173\linewidth}
        \centering
        \fbox{\includegraphics[width=\linewidth]{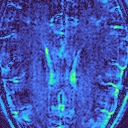}}
        \caption{\small \textbf{Ours}}
        \label{fig:diff_mdsr}
    \end{subfigure}
    
    \caption{Qualitative comparison of PD-guided T2 SR error maps for the region shown on the GT image. Subsequent panels show absolute error heatmaps for baselines and our model.}
    \label{fig:qualitative_diff_maps}
\end{figure}

\section{Experiments and Results}
\label{sec:experiments}

We evaluate CSC-MAMBA on the IXI brain MRI dataset\footnote{\url{https://brain-development.org/ixi-dataset/}}, comprising 576 multi-contrast scans (T1WI, T2WI, and PDWI) from healthy subjects acquired across multiple clinical sites. For our multi-contrast SR experiments---specifically targeting T2-weighted imaging with PD images as the reference modality---the dataset is partitioned subject-wise to prevent data leakage, yielding 23,000 training slices, 5,700 validation slices, and 2,930 test slices. LR T2 inputs are generated via $k$-space truncation at scale factors of $\times2$ and $\times4$ followed by zero-padding to match the spatial dimensions of the HR PD reference. The network is trained on an NVIDIA RTX 4060 Ti GPU using the Adam optimizer. The learning rate is initialized at $1 \times 10^{-3}$ for the first 50 epochs and subsequently reduced to $1 \times 10^{-4}$ for the remaining 50 epochs. All reported inference times represent the average processing time across the entire test set evaluated on this same GPU.

We compare CSC-MAMBA with six methods spanning several categories, including convolutional dictionary (CD-DPE~\cite{gu2026cddpe}), diffusion models (DiffMSR~\cite{li2024rethinking}), transformers (McMRSR~\cite{li2022transformer}), Mamba-based models (DDHMamba~\cite{ddh2026mamba}), deep unfolding networks (HocMRI~\cite{shen2026unfolding}), and decomposition-based frameworks (McVarNet~\cite{lei2023decomposition}). Reconstruction quality is evaluated using peak signal-to-noise ratio (PSNR) and structural similarity (SSIM), with higher values indicating better reconstruction fidelity and structural preservation, respectively. All methods are evaluated on the same test set using consistent preprocessing and evaluation protocols.

Table~\ref{tab:quantitative_comparison} summarizes the quantitative results and inference times at $\times2$ and $\times4$ scales. CSC-MAMBA achieves state-of-the-art performance, surpassing the strongest baseline (CD-DPE) by 0.49~dB at $\times2$ and 0.62~dB at $\times4$ in PSNR, while cutting inference latency nearly in half (52~ms vs. 98~ms). These quantitative gains, corroborated by the visual comparisons in Fig.~\ref{fig:qualitative_diff_maps}, demonstrate the effectiveness of combining interpretable, model-guided cross-contrast fusion with Mamba-based long-range dependency modeling under varying degradation levels.



\section{Conclusion}
In this paper, we proposed \textbf{CSC-MAMBA}, a hybrid framework for multi-contrast MRI super-resolution that resolves the trade-off between principled feature fusion and global representation capacity. By fusing common structural priors via unfolded convolutional sparse coding and refining contrast-specific representations with a Mamba backbone, CSC-MAMBA inherently couples the theoretical foundations of model-based methods with long-range spatial modeling. Experiments on the IXI dataset confirm that our approach achieves state-of-the-art reconstruction performance while maintaining computational efficiency, highlighting the clinical potential of integrating explicit sparse priors with state-space architectures.



\end{document}